\documentclass{article} 
\usepackage{nips12submit_e,times}

\title{Barnes-Hut-SNE}

\author{
Laurens van der Maaten\\
Pattern Recognition and Bioinformatics Group, Delft University of Technology\\
Mekelweg 4, 2628 CD Delft, The Netherlands\\
\texttt{lvdmaaten@gmail.com} \\
}

\usepackage{algorithm}
\usepackage{algorithmic}
\usepackage{graphicx}
\usepackage{amsmath}
\usepackage{amssymb}
\usepackage{color}
\usepackage{colortbl}
\usepackage{tabularx}
\usepackage{subfigure}
\usepackage{arydshln}
\usepackage{url}
\usepackage{multirow}
\usepackage{epstopdf}
\usepackage{wrapfig}

\nipsfinalcopy 

\begin{document}

\maketitle

\begin{abstract}
The paper presents an $\mathcal{O}(N \log N)$-implementation of t-SNE --- an embedding technique that is commonly used for the visualization of high-dimensional data in scatter plots and that normally runs in $\mathcal{O}(N^2)$. The new implementation uses vantage-point trees to compute sparse pairwise similarities between the input data objects, and it uses a variant of the Barnes-Hut algorithm to approximate the forces between the corresponding points in the embedding. Our experiments show that the new algorithm, called Barnes-Hut-SNE, leads to substantial computational advantages over standard t-SNE, and that it makes it possible to learn embeddings of data sets with millions of objects.
\end{abstract}

\section{Introduction}
Data-visualization techniques are an essential tool for any data analyst, as they allow the analyst to visually explore the data and generate hypotheses. One of the key limitations of traditional visualization techniques such as histograms, scatter plots, and parallel coordinate plots (see, \emph{e.g.}, \cite{heer10} for an overview) is that they only facilitate the visualization of one or a few data variables at a time. In order to get an idea of the structure of all variables in the data, it is therefore necessary to perform an automatic analysis of the data before making the visualization, for instance, by learning a low-dimensional embedding of the data. In such an embedding, each data object is represented by a low-dimensional point in such a way, that nearby points correspond to similar data objects and that distant points correspond to dissimilar data objects. The low-dimensional embedding can readily be visualized in, \emph{e.g.}, a scatter plot or a parallel coordinate plot.

A plethora of embedding techniques have been proposed over the last decade, \emph{e.g.}, \cite{carreira10,lawrence11,roweis00,tenenbaum00,vandermaaten08d,venna10}. For creating two- or three-dimensional embeddings that can be readily visualized in a scatter plot, a family of techniques based on \emph{stochastic neighbor embedding} (SNE; \cite{hinton03}) has recently become very popular. These techniques compute an $N \!\times\! N$ similarity matrix in both the original data space and in the low-dimensional embedding space; the similarities take the form of a probability distribution over pairs of points in which high probabilities correspond to similar objects or points. The probabilities are generally defined as normalized Gaussian or Student-t kernels, which makes that SNE focuses on preserving \emph{local} data structure. The embedding is learned by minimizing the Kullback-Leibler divergence between the probability distributions in the original data space and the embedding space with respect to the locations of the points in the embedding. As the resulting cost function is non-convex, this minimization is typically performed using first- or second-order gradient-descent techniques \cite{carreira10,hinton03,vladymyrov12}. The gradient of the Kullback-Leibler divergence may be interpreted as an $N$-body system in which all of the $N$ points exert forces on each other.

One of the key limitations of SNE (and of its variants) is that its computational and memory complexity scales quadratically in the number of data objects $N$. In practice, this limits the applicability of SNE to data sets with only a few thousand points. To visualize larger data sets, landmark implementations of SNE may be used \cite{vandermaaten08d}, but this is hardly a satisfactory solution.

In this paper, we develop a new algorithm for (t-)SNE that requires only $\mathcal{O}(N \log N)$ computation and $\mathcal{O}(N)$ memory. Our new algorithm computes a sparse approximation of the similarities between the original data objects using vantage-point trees \cite{yianilos93}, and subsequently, approximates the forces between the points in the embedding using a Barnes-Hut algorithm \cite{barnes86} --- an algorithm commonly used by astronomers to perform $N$-body simulations. The Barnes-Hut algorithm reduces the number of pairwise forces that needs to be computed by exploiting the fact that the forces exerted by a group of points on a point that is relatively far away are all very similar. 

\section{Related work}
\label{Related work}
A large body of previous work has focused on decreasing the computational complexity of algorithms that scale quadratically in the amount of data when implemented naively. Most of these studies focus on speeding up nearest-neighbor searches using space-partitioning (metric) trees (\emph{e.g.}, B-trees \cite{bayer72}, cover trees \cite{beygelzimer06}, and vantage-point trees \cite{yianilos93}) or using locality sensitive hashing approaches (\emph{e.g.}, \cite{indyk98,weiss08}). Motivated by their strong performance reported in earlier work in \cite{liu04}, we opt to use metric trees to approximate the similarities of the input objects in our algorithm.

Several prior studies have also developed algorithms to speed up $N$-body computations. Most prominently, \cite{gray01,gray03} developed a dual-tree algorithm that is similar in spirit to the Barnes-Hut algorithm we use in this work. The dual-tree algorithm does not consider interactions between single points and groups of points like the Barnes-Hut algorithm, but it only considers interactions between groups of points. In preliminary experiments (see appendix), we found the dual-tree and Barnes-Hut algorithms to perform on par when used in the context of t-SNE --- we opt for the Barnes-Hut algorithm here because it is conceptually simpler. Prior work \cite{defreitas06} has also used the fast Gaussian transform \cite{greengard87,yang03} (a special case of a fast multipole method \cite{rokhlin85}) to speed up the computation of Gaussian $N$-body interactions. Since in t-SNE, the forces exerted on the bodies are non-Gaussian, such an approach cannot readily be applied here.

\section{t-Distributed Stochastic Neighbor Embedding}
\label{t-Distributed Stochastic Neighbor Embedding}
t-Distributed Stochastic Neighbor Embedding (t-SNE) minimizes the divergence between two distributions: a distribution that measures pairwise similarities between the original data objects and a distribution that measures pairwise similarities between the corresponding points in the embedding. Suppose we are given a data set of objects $\mathcal{D} = \{\mathbf{x}_1, \mathbf{x}_2, \dots, \mathbf{x}_N\}$ and a function $d(\mathbf{x}_i, \mathbf{x}_j)$ that computes a distance between a pair of objects, \emph{e.g.}, their Euclidean distance. Our aim is to learn an $s$-dimensional embedding in which each object is represented by a point, $\mathcal{E} = \{\mathbf{y}_1, \mathbf{y}_2, \dots, \mathbf{y}_N\}$ with $\mathbf{y}_i \in \mathbb{R}^s$. To this end, t-SNE defines joint probabilities $p_{ij}$ that measure the pairwise similarity between objects $\mathbf{x}_i$ and $\mathbf{x}_j$ by symmetrizing two conditional probabilities as follows:
\begin{align}
p_{j|i} &= \frac{\exp(-d(\mathbf{x}_i, \mathbf{x}_j)^2 / 2\sigma_i^2)}{\sum_{k \neq i} \exp(-d(\mathbf{x}_i, \mathbf{x}_k)^2 / 2\sigma_i^2)}, &p_{i|i} = 0\\
p_{ij} &= \frac{p_{j|i} + p_{i|j}}{2N}.
\end{align}
Herein, the bandwidth of the Gaussian kernels $\sigma_i$ is set such that the perplexity of the conditional distribution $P_i$ equals a predefined perplexity $u$. The optimal value of $\sigma_i$ varies per object, and is found using a simple binary search; see \cite{hinton03} for details. A heavy-tailed distribution is used to measure the similarity $q_{ij}$ between the two corresponding points $\mathbf{y}_i$ and $\mathbf{y}_j$ in the embedding:
\begin{align}
q_{ij} = \frac{(1 + \lVert \mathbf{y}_i - \mathbf{y}_j\rVert^2)^{-1}}{\sum_{k \neq l} (1 + \lVert \mathbf{y}_k - \mathbf{y}_l)\rVert^2)^{-1}},~~~~~~~~~~~~~~~~~~~~~~~~~~~~~~~~~&q_{ii} = 0.
\end{align}
In the embedding, a normalized Student-t kernel is used to measure similarities rather than a normalized Gaussian kernel to account for the difference in volume between high- and low-dimensional spaces \cite{vandermaaten08d}. The locations of the embedding points $\mathbf{y}_i$ are learned by minimizing the Kullback-Leibler divergence between the joint distributions $P$ and $Q$:
\begin{equation}
C(\mathcal{E}) = KL(P||Q) = \sum_{i \neq j} p_{ij} \log \frac{p_{ij}}{q_{ij}}.\label{eq:cost}
\end{equation}
This cost function is non-convex; it is typically minimized by descending along the gradient:
\begin{equation}
\frac{\partial C}{\partial \mathbf{y}_i} = 4 \sum_{j \neq i} (p_{ij} - q_{ij}) q_{ij} Z (\mathbf{y}_i - \mathbf{y}_j),
\end{equation}
where we defined the normalization term $Z = \sum_{k \neq l} (1 + \lVert \mathbf{y}_k - \mathbf{y}_l)\rVert^2)^{-1}$. The evaluation of both joint distributions $P$ and $Q$ is $\mathcal{O}(N^2)$, because their respective normalization terms sum over all $N^2$ pairs of points. Since t-SNE scales quadratically in the number of objects $N$, its applicability is limited to data sets with only a few thousand data objects; beyond that, learning becomes very slow.

\section{Barnes-Hut-SNE}
\label{Barnes-Hut-SNE}
Barnes-Hut-SNE uses metric trees to approximate $P$ by a sparse distribution in which only $\mathcal{O}(uN)$ values are non-zero, and approximates the gradients $\frac{\partial C}{\partial \mathbf{y}_i}$ using a Barnes-Hut algorithm. 

\subsection{Approximating Input Similarities} 
As the input similarities are computed using a (normalized) Gaussian kernel, probabilities $p_{ij}$ corresponding to dissimilar input objects $i$ and $j$ are (nearly) infinitesimal. Therefore, we can use a sparse approximation to the probabilities $p_{ij}$ without a substantial negative effect on the quality of the final embeddings. In particular, we compute the sparse approximation by finding the $\lfloor 3u \rfloor$ nearest neighbors of each of the $N$ data objects, and redefining the pairwise similarities $p_{ij}$ as:
\begin{equation}
p_{j|i} = \left\{ \begin{array}{rl}
 \frac{\exp(-d(\mathbf{x}_i, \mathbf{x}_j)^2 / 2\sigma_i^2)}{\sum_{k \in \mathcal{N}_i} \exp(-d(\mathbf{x}_i, \mathbf{x}_k)^2 / 2\sigma_i^2)}, &\mbox{~~~~~if $j \in \mathcal{N}_i$} \\
  0, &\mbox{~~~~~otherwise}
       \end{array} \right.
\end{equation}
\begin{equation}
p_{ij} = \frac{p_{j|i} + p_{i|j}}{2N}.
\end{equation}
Herein, $\mathcal{N}_i$ represents the set of the $\lfloor 3u \rfloor$ nearest neighbors of $\mathbf{x}_i$, and $\sigma_i$ is set such that the perplexity of the conditional distribution equals $u$. The nearest neighbor sets are found in $\mathcal{O}(uN \log N)$ time by building a vantage-point tree on the data set.

\textbf{Vantage-point tree.} In a vantage-point tree, each node stores a data object and the radius of a (hyper)ball that is centered on this object \cite{yianilos93}. All non-leaf nodes have two children: data objects that are located \emph{inside} the ball are stored under the left child of the node, whereas data objects that are located \emph{outside} the ball are stored under the right child. The tree is constructed by presenting the data objects one-by-one, traversing the tree based on whether the current data object lies inside or outside a ball, and creating a new leaf node in which the object is stored. The radius of the new leaf node is set to the median distance between its object and all other objects that lie inside the ball represented by its parent node. To construct a vantage-point tree, the objects need not necessarily be points in a high-dimensional feature space; the availability of a metric $d(\mathbf{x}_i, \mathbf{x}_j)$ suffices. (In our experiments, however, we use $\mathbf{x}_i \in \mathbb{R}^D$ and $d(\mathbf{x}_i, \mathbf{x}_j) = \lVert \mathbf{x}_i - \mathbf{x}_j\rVert$.)

A nearest-neighbor search is performed using a depth-first search on the tree that computes the distance of the objects stored in the nodes to the target object, whilst maintaining i) a list of the current nearest neighbors and ii) the distance $\tau$ to the furthest nearest neighbor in the current neighbor list. The value of $\tau$ determines whether or not a node should be explored: if there can still be objects inside the ball whose distance to the target object is smaller than $\tau$, the left node is searched, and if there can still be objects outside the ball whose distance to the target object is smaller than $\tau$, the right node is searched. The order in which children are searched depends on whether or not the target object lies inside or outside the current node ball: the left child is examined first if the object lies inside the ball, because the odds are that the nearest neighbors of the target object are also located inside the ball. The right child is searched first whenever the target object lies outside of the ball.

\subsection{Approximating t-SNE Gradients}
To approximate the t-SNE gradient, we start by splitting the gradient into two parts as follows:
\begin{equation}
\frac{\partial C}{\partial \mathbf{y}_i} = 4 (F_{attr} - F_{rep}) = 4 \left(\sum_{j \neq i} p_{ij} q_{ij} Z (\mathbf{y}_i - \mathbf{y}_j) - \sum_{j \neq i} q_{ij}^2 Z (\mathbf{y}_i - \mathbf{y}_j)\right),
\end{equation}
where $F_{attr}$ denotes the sum of all attractive forces (the left sum), whereas $F_{rep}$ denotes the sum of all repulsive forces (the right sum). Computing the sum of all attractive forces, $F_{attr}$, is computationally efficient; it can be done by summing over all non-zero elements of the sparse distribution $P$ in $\mathcal{O}(uN)$. (Note that the term $q_{ij} Z = (1 + \lVert \mathbf{y}_i - \mathbf{y}_j\rVert^2)^{-1}$ can be computed in $\mathcal{O}(1)$.) However, a naive computation of the sum of all repulsive forces, $F_{rep}$, is $\mathcal{O}(N^2)$. We now develop a Barnes-Hut algorithm to approximate $F_{rep}$ efficiently in $\mathcal{O}(N \log N)$.

Consider three points $\mathbf{y}_i$, $\mathbf{y}_j$, and $\mathbf{y}_k$ with $\lVert \mathbf{y}_i - \mathbf{y}_j\rVert \!\approx\! \lVert \mathbf{y}_i - \mathbf{y}_k\rVert \!\gg\! \lVert \mathbf{y}_j - \mathbf{y}_k\rVert$. In this situation, the contributions of $\mathbf{y}_j$ and $\mathbf{y}_k$ to $F_{rep}$ will be roughly equal. The Barnes-Hut algorithm \cite{barnes86} exploits this by i) constructing a quadtree on the current embedding, ii) traversing the quadtree using a depth-first search, and iii) at every node in the quadtree, deciding whether the corresponding cell can be used as a ``summary'' for the gradient contributions of all points in that cell.

\begin{wrapfigure}{rt}{.5\textwidth}
\centering
\includegraphics[width=.5\textwidth]{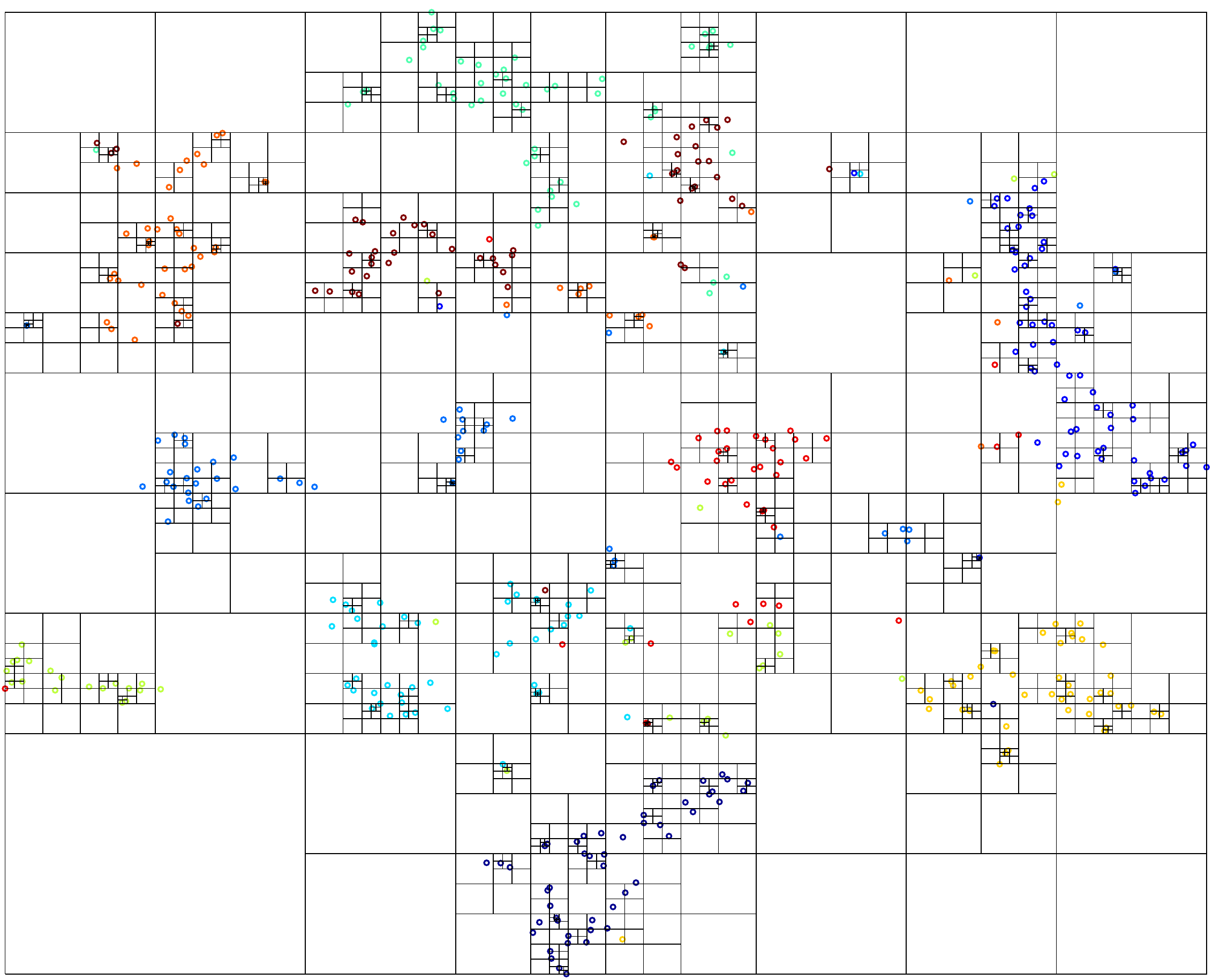}
\caption{Quadtree constructed on a two-dimensional t-SNE embedding of $500$ MNIST digits (the colors of the points correspond to the digit classes). Note how the quadtree adapts to the local point density in the embedding.}
\label{fig:quadtree}
\end{wrapfigure}

\textbf{Quadtree.} A quadtree is a tree in which each node represents a rectangular \emph{cell} with a particular center, width, and height. Non-leaf nodes have four children that split up the cell into four smaller cells (quadrants) that lie ``northwest'', ``northeast'', ``southwest'', and ``southeast'' of the center of the parent node (see Figure~\ref{fig:quadtree} for an illustration). Leaf nodes represent cells that contain at most one point of the embedding; the root node represents the cell that contains the complete embedding. In each node, we store the center-of-mass of the embedding points that are located inside the corresponding cell, $\mathbf{y}_{cell}$, and the total number of points that lie inside the cell, $N_{cell}$. A quadtree has $\mathcal{O}(N)$ nodes and can be constructed in $\mathcal{O}(N)$ time by inserting the points one-by-one, splitting a leaf node whenever a second point is inserted in its cell, and updating $\mathbf{y}_{cell}$ and $N_{cell}$ of all visited nodes. 

\textbf{Approximating the gradient.} To approximate the repulsive part of the gradient, $F_{rep}$, we note that if a cell is sufficiently small and sufficiently far away from point $\mathbf{y}_i$, the contributions $q_{ij}^2 Z (\mathbf{y}_i - \mathbf{y}_j)$ to $F_{rep}$ will be roughly similar for all points $\mathbf{y}_j$ inside that cell. We can, therefore, approximate these contributions by $N_{cell} q_{i, cell}^2 Z (\mathbf{y}_i - \mathbf{y}_{cell})$, where 
we define $q_{i, cell}Z = (1 + \Vert \mathbf{y}_i - \mathbf{y}_{cell}\rVert^2)^{-1}$. We first approximate $F_{rep} Z = q_{ij}^2 Z^2 (\mathbf{y}_i - \mathbf{y}_j)$ by performing a depth-first search on the quadtree, assessing at each node whether or not that node may be used as a ``summary'' for all the embedding points that are located in the corresponding cell. During this search, we construct an estimate of $Z = \sum_{i \neq j} (1 + \lVert \mathbf{y}_i - \mathbf{y}_j\rVert^2)^{-1}$ in the same way. The two approximations thus obtained are then used to compute $F_{rep}$ via $F_{rep} = \frac{F_{rep}Z}{Z}$.

We use the condition proposed by \cite{barnes86} to decide whether a cell may be used as a ``summary'' for all points in that cell. The condition compares the distance of the cell to the target point with its size:
\begin{equation}
\lVert \mathbf{y}_i - \mathbf{y}_{cell} \rVert^2 / r_{cell} < \theta,\label{eq:condition}
\end{equation}
where $r_{cell}$ represents the length of the diagonal of the cell under consideration and $\theta$ is a threshold that trades off speed and accuracy (higher values of $\theta$ lead to poorer approximations). In preliminary experiments, we also explored various other conditions that take into account the rapid decay of the Student-t tail, but we did not find to lead these alternative conditions to lead to a better accuracy-speed trade-off. (The problem of more complex conditions is that they require expensive computations at each cell. By contrast, the condition in Equation \ref{eq:condition} can be evaluated very rapidly.) 

\textbf{Dual-tree algorithms.} Whilst the Barnes-Hut algorithm considers \emph{point-cell} interactions, further speed-ups may be obtained by computing only \emph{cell-cell} interactions. This can be done using a dual-tree algorithm \cite{gray01} that simultaneously traverses the quadtree twice, and for every pair of nodes decides whether the interaction between the corresponding cells can be used as ``summary'' for the interactions between all points inside these two cells. Perhaps surprisingly, we did find such an approach to perform on par with the Barnes-Hut algorithm in preliminary experiments. The computational advantages of the dual-tree algorithm evaporate because after computing an interaction between two cells, one still needs to determine to which set of points the interaction applies. This can be done by searching the cell or by storing a list of children in each node during tree construction. Both these approaches are computationally costly. (It should be noted that the dual-tree algorithm is, however, much faster in approximating the value of the t-SNE cost function.) The results of our experiments with dual-tree algorithms are presented in the appendix.

\section{Experiments}
\label{Experiments}
We performed experiments on four large data sets to evaluate the performance of Barnes-Hut-SNE. Code for our algorithm is available from \url{http://homepage.tudelft.nl/19j49/tsne}.

\textbf{Data sets.} We performed experiments on four data sets: i) the MNIST data set, ii) the CIFAR-10 data set, iii) the NORB data set, and iv) the TIMIT data set. The MNIST data set contains $N\!=\! 70,000$ grayscale handwritten digit images of size $D\!=\! 28\!\times\! 28 \!=\! 784$ pixels, each of which corresponds to one of ten classes. The CIFAR-10 data set \cite{krizhevsky09} is an annotated subset of the 80 million tiny images data set \cite{torralba08} that contains $N\!=\!70,000$ RGB images of size $32\!\times\! 32$ pixels, leading to a $D\!=\! 32\!\times\! 32 \!\times\! 3 \!=\! 3,072$-dimensional input objects; each image corresponds to one of ten classes. The (small) NORB data set \cite{lecun04} contains grayscale images of toys from five different classes, rendered on a uniform background under $6$ lighting conditions, $9$ elevations ($30$ to $70$ degrees every 5 degrees), and $18$ azimuths ($0$ to $340$ every $20$ degrees). All images contain $D\!=\! 96\!\times\!96\!=\! 9,216$ pixels. The TIMIT data set contains speech data from which MFCC, delta, and delta-delta features were extracted, leading to $D\!=\!39$-dimensional features \cite{sha07}; each frame in the data has one of $39$ phone labels. We used the TIMIT training set of $N\! =\! 1,105,455$ frames in our experiments.

\begin{figure*}[t]
\centering
\includegraphics[width=\textwidth]{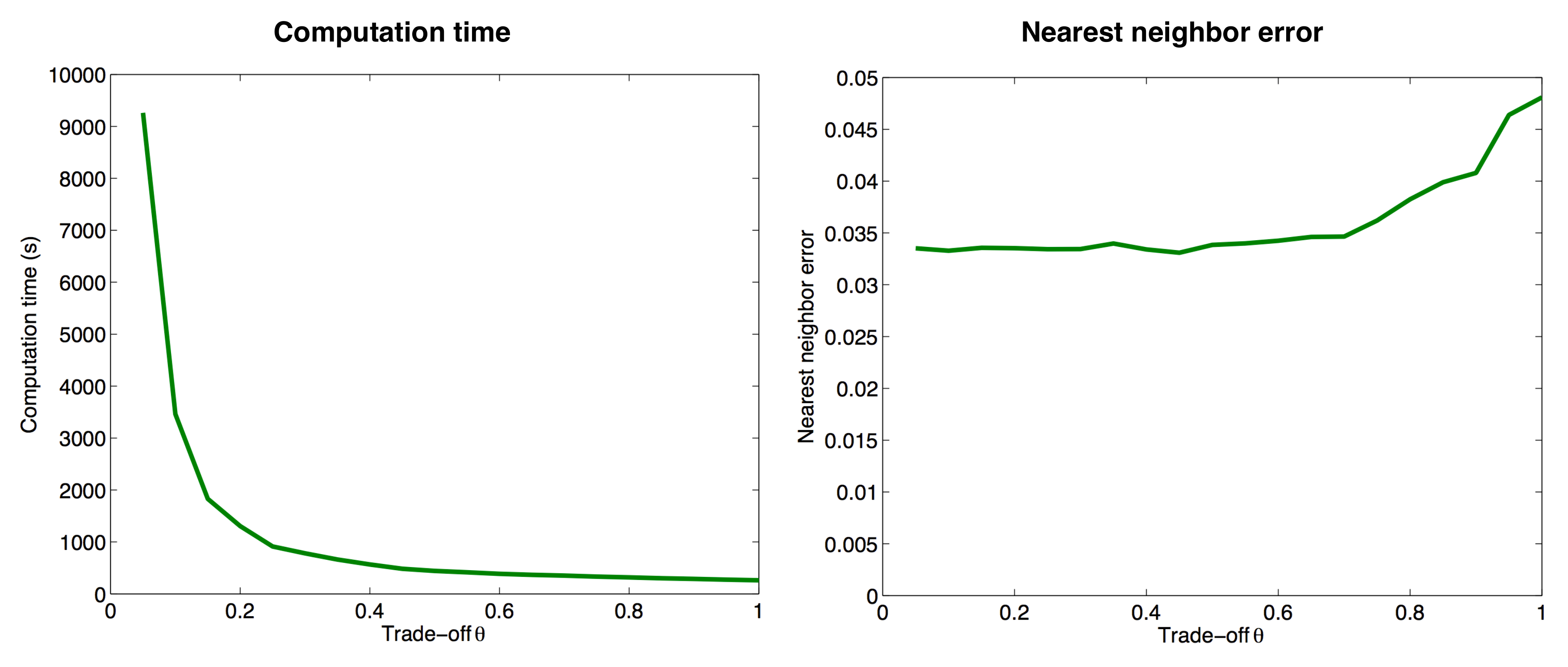}
\caption{Computation time (in seconds) required to embed $70,000$ MNIST digits using Barnes-Hut-SNE (left) and the $1$-nearest neighbor errors of the corresponding embeddings (right) as a function of the trade-off parameter $\theta$.}
\label{fig:mnist_theta}
\end{figure*}

\textbf{Experimental setup.} In all our experiments, we follow the experimental setup of \cite{vandermaaten08d} as closely as possible. In particular, we initialize the embedding points by sampling from a Gaussian with a variance of $10^{-4}$. We run a gradient-descent optimizer for $1,000$ iterations, setting the initial step size to $200$; the step size is updated during the optimization use the scheme of \cite{jacobs88}. We use an additional momentum term that has weight $0.5$ during the first $250$ iterations, and $0.8$ afterwards. The perplexity $u$ is fixed to $30$. Following \cite{vandermaaten08d}, all data sets with a dimensionality $D$ larger than $50$ were preprocessed using PCA to reduce their dimensionality to $50$.

During the first $250$ learning iterations, we multiplied all $p_{ij}$-values by a user-defined constant $\alpha \! > \! 1$. As explained in \cite{vandermaaten08d}, this trick enables t-SNE to find a better global structure in the early stages of the optimization. In preliminary experiments, we found that this trick becomes increasingly important to obtain good embeddings when the data set size increases, as it becomes harder for the optimization to find a good global structure when there are more points in the embedding because there is less space for clusters to move around. In our experiments, we fix $\alpha \! = \! 12$ (by contrast, \cite{vandermaaten08d} used $\alpha \! = \! 4$).

We present the results of three sets of experiments. In the first experiment, we investigate the effect of the trade-off parameter $\theta$ on the speed and the quality of embeddings produced by Barnes-Hut-SNE on the MNIST data set. In the second experiment, we investigate the computation time required to run Barnes-Hut-SNE as a function of the number of data objects $N$ (also on the MNIST data set). In the third experiment, we construct and visualize embeddings of all four data sets.

\textbf{Results.} 
Figure~\ref{fig:mnist_theta} presents the results of an experiment in which we varied the speed-accuracy trade-off parameter $\theta$ used to learn the embedding. The figure shows the computation time required to construct embeddings of all $70,000$ MNIST digit images, as well as the $1$-nearest neighbor error (computed based on the digit labels) of the corresponding embeddings. The results presented in the figure show that the trade-off parameter $\theta$ may be increased to a value of approximately $0.5$ without negatively affecting the quality of the embedding. At the same time, increasing the value of $\theta$ to $0.5$ leads to very substantial improvements in terms of the amount of computation required: the time required to embed all $70,000$ MNIST digits is reduced to just $645$ seconds when $\theta\! = \! 0.5$. (Note that the special case $\theta\! = \! 0$ corresponds to standard t-SNE \cite{vandermaaten08d}; we did not run an experiment with $\theta\! = \! 0$ because standard t-SNE would take days to complete on the full MNIST data set.) 

\begin{figure*}[t]
\centering
\includegraphics[width=\textwidth]{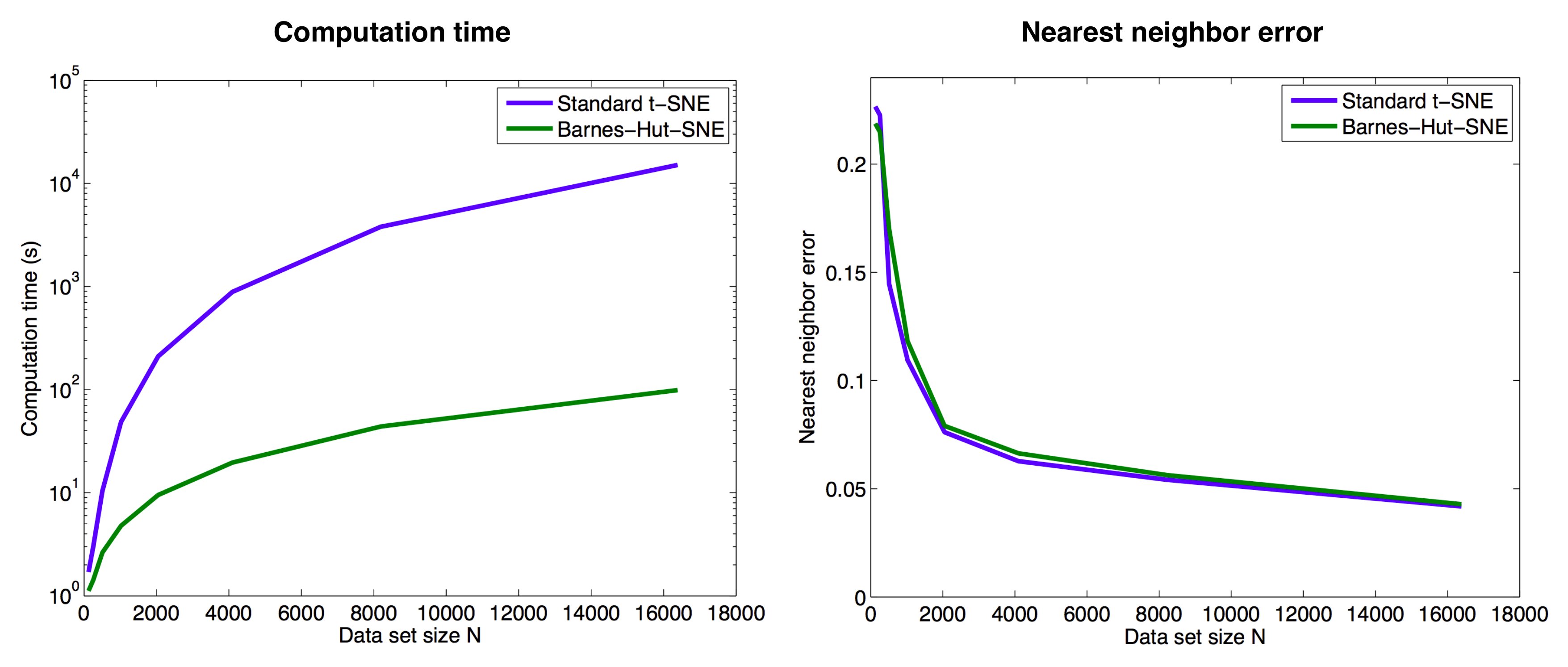}
\caption{Compution time (in seconds) required to embed MNIST digits (left) and the $1$-nearest neighbor errors of the corresponding embeddings (right) as a function of data set size $N$ for both standard t-SNE and Barnes-Hut-SNE. Note that the required computation time, which is shown on the $y$-axis of the left figure, is plotted on a logarithmic scale.}
\label{fig:mnist_perf}
\end{figure*}

In Figure~\ref{fig:mnist_perf}, we compare standard t-SNE and Barnes-Hut-SNE in terms of i) the computation time required for the embedding of MNIST digit images as a function of the data set size $N$ and ii) the $1$-nearest neighbor errors of the corresponding embeddings.
(Note that the $y$-axis of the left figure, which represents the required computation time in seconds, uses a logarithmic scale.) In the experiments, we fixed the parameter $\theta$ that trades off speed and accuracy to $0.5$. The results presented in the figure show that Barnes-Hut-SNE is orders of magnitude faster than standard t-SNE, whilst the difference in quality of the constructed embeddings (which is measured by the nearest-neighbor errors) is negligible. Most prominently, the computational advantages of Barnes-Hut-SNE rapidly increase as the number of objects in the data set $N$ increases.

\begin{figure*}[t]
\centering
\includegraphics[width=\textwidth]{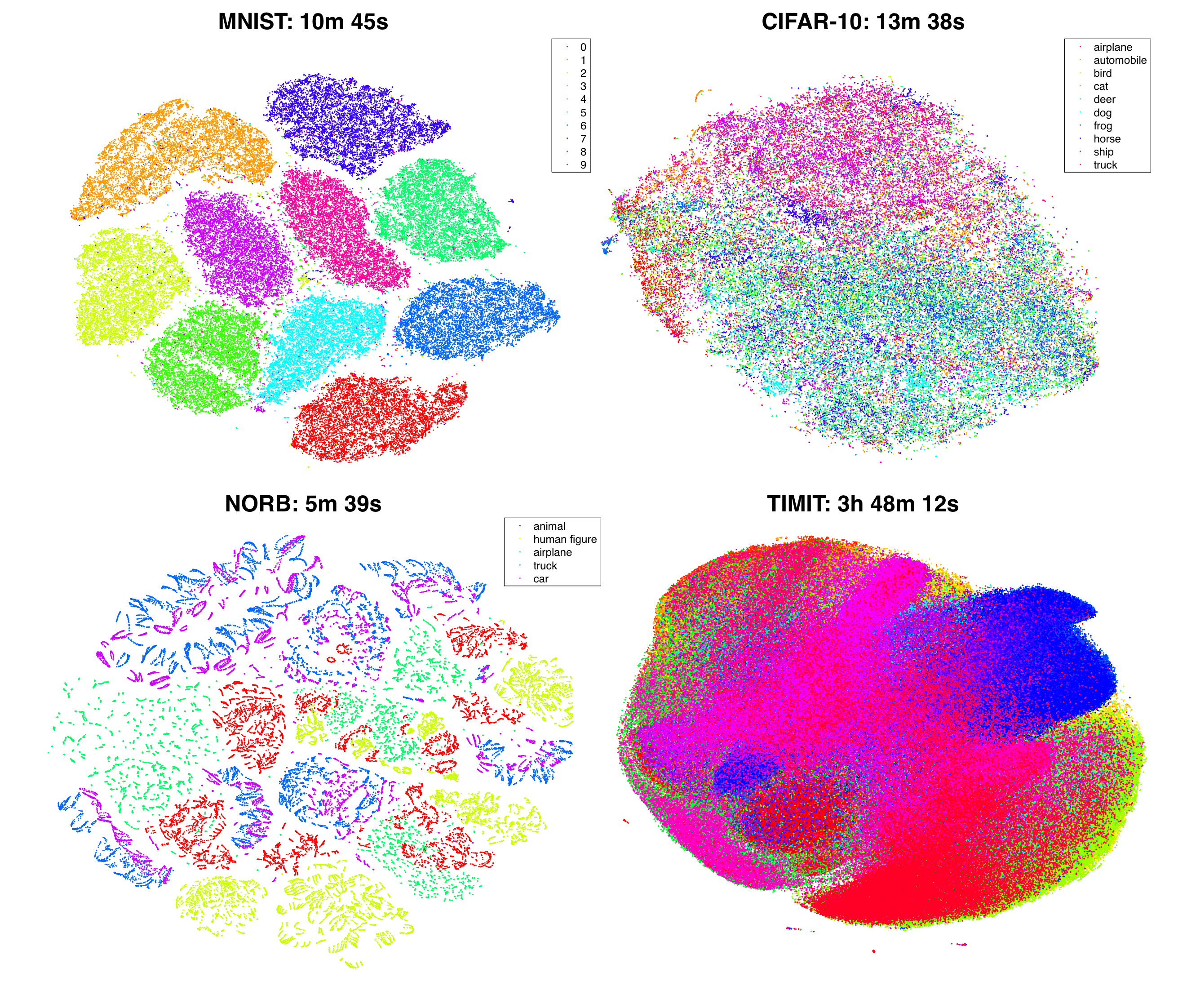}
\caption{Barnes-Hut-SNE visualizations of four data sets: MNIST handwritten digits (top-left), CIFAR-10 tiny images (top-right), NORB object images (bottom-left), and TIMIT speech frames (bottom-right). The colors of the point indicate the classes of the corresponding objects. The titles of the figures indicate the computation time that was used to construct the corresponding embeddings. Figure best viewed in color.}
\label{fig:all_scatter_plots}
\end{figure*}

Figure~\ref{fig:all_scatter_plots} presents embeddings of all four data sets constructed using Barnes-Hut-SNE. The colors of the points indicate the classes of the corresponding objects; the titles of the plots indicate the computation time that was used to construct the corresponding embeddings. As before, we fixed $\theta\! = \! 0.5$ in all four experiments. The results in the figure shows that Barnes-Hut-SNE can construct high-quality embeddings of, \emph{e.g.}, the $70,000$ MNIST handwritten digit images in just over $10$ minutes. (Although our MNIST embedding contains many more points, it may be compared with that in \cite{vandermaaten08d}. Visually, the structure of the two embeddings is very similar.) The results also show that Barnes-Hut-SNE makes it practical to embed data sets with more than a million data points: the TIMIT embedding shows all $1,105,455$ data points, and was constructed in less than four hours.

\begin{figure*}[t]
\centering
\includegraphics[width=\textwidth]{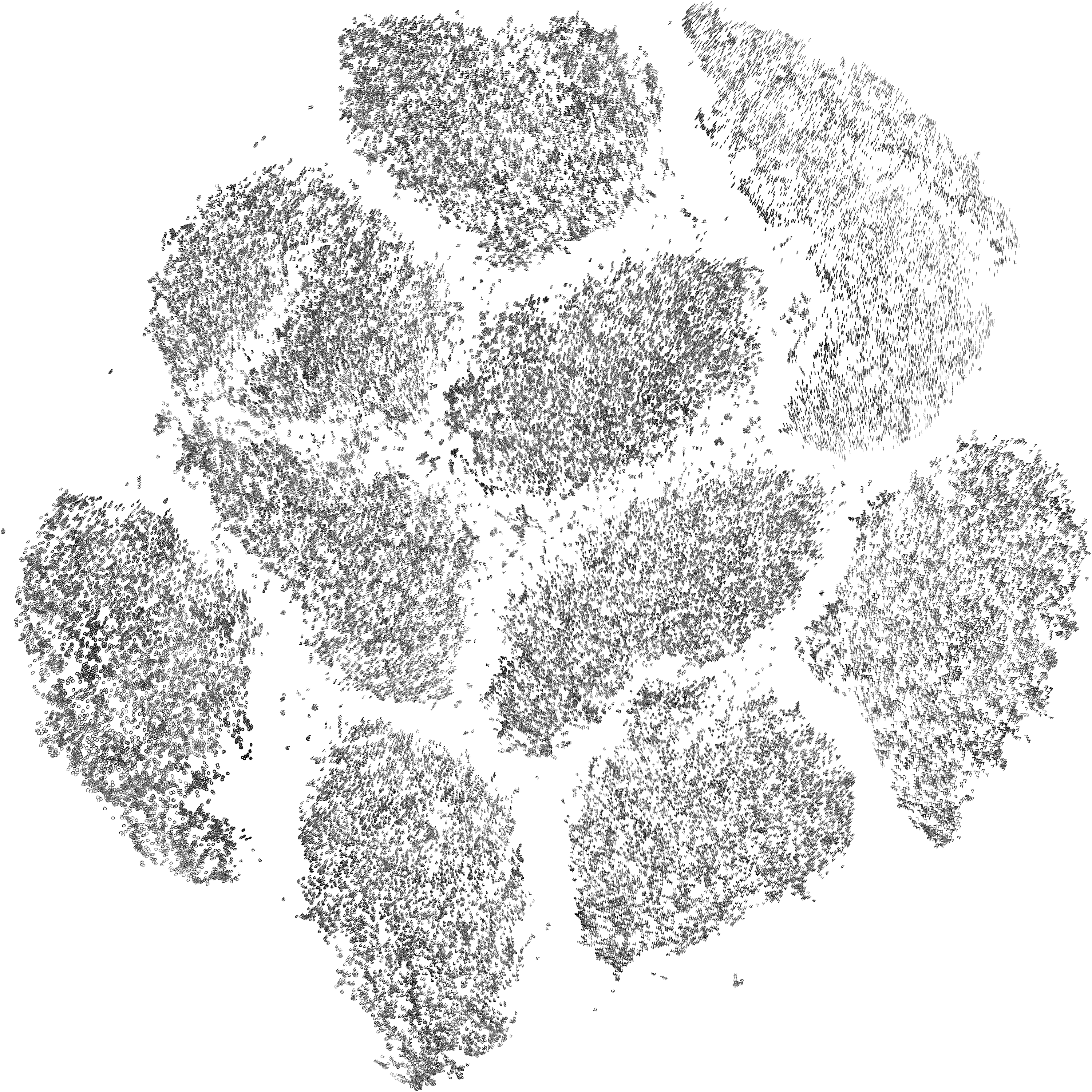}
\caption{Barnes-Hut-SNE visualization of all $70,000$ MNIST handwritten digit images (constructed in 10 minutes and 45 seconds). Zoom in on the visualization for more detailed views.}
\label{fig:mnist_map}
\end{figure*}

A version of the MNIST embedding in which the original digit images are shown is presented in Figure~\ref{fig:mnist_map}. The results show that, like standard t-SNE, Barnes-Hut-SNE is very good at preserving local structure of the data in the embedding: for instance, the visualization clearly shows that orientation is one of the main sources of variation within the cluster of ones. 

\section{Conclusion and Future Work}
\label{Discussion}
We presented a new t-SNE algorithm \cite{vandermaaten08d}, called Barnes-Hut-SNE, that i) constructs a sparse approximation to the similarities between input objects using vantage-point trees, and ii) approximates the t-SNE gradient using a variant of the Barnes-Hut algorithm. The new algorithm runs in $\mathcal{O}(N \log N)$ rather than $\mathcal{O}(N^2)$, and requires only $\mathcal{O}(N)$ memory. Our experimental evaluation of Barnes-Hut-SNE shows that it is substantially faster than standard t-SNE, and that it facilitates the visualization of data sets with millions of data objects in scatter plots. 

A drawback of Barnes-Hut-SNE is that it does not provide any error bounds \cite{salmon94}. Indeed, there exist alternative algorithms that do provide such error bounds (\emph{e.g.}, \cite{warren93}); we aim to explore these alternatives in future work to see whether they can be used to bound the error made in our t-SNE gradient computations, and to bound the error in the final embedding. Another limitation of Barnes-Hut-SNE is that it can only be used to embed data in two or three dimensions. Generalizations to higher dimensions are infeasible because the size of the tree grows exponentially in the dimensionality of the embedding space. Having said that, this limitation is not very severe since t-SNE is mainly used for visualization (\emph{i.e.} for embedding in two or three dimensions). Moreover, it is relatively straightforward to replace the quadtree by metric trees that scale better to high-dimensional spaces.

In future work, we plan to further scale up our algorithm by developing parallelized implementations that can run on data sets that are too large to be fully stored in memory. We also aim to investigate the effect of varying the value of $\theta$ during the optimization. In addition, we plan to explore to what extent adapted versions of our algorithm (that use metric trees instead of quadtrees) can be used to speed up techniques for relational embedding (\emph{e.g.}, \cite{bordes11,paccanaro01}).

\section*{Acknowledgments}
The author is supported by EU-FP7 Social Signal Processing (SSPNet) and by the Netherlands Institue for Advanced Study (NIAS). The author thanks Geoffrey Hinton for many helpful discussions, and two anonymous reviewers for their helpful comments.

\small{
\bibliography{references}
\bibliographystyle{plain}
}

\newpage
\appendix
\section{Experiments with Dual-Tree t-SNE}
\label{Experiments with Dual-Tree t-SNE}
We also performed experiments with a dual-tree implementation \cite{gray01} of t-SNE. Dual-tree t-SNE differs from Barnes-Hut-SNE in that it considers only \emph{cell-cell} instead of \emph{point-cell} interactions. It simultaneously traverses the quadtree twice, and decides for each pair of nodes whether the interaction between these nodes can be used as a ``summary'' for all points in the cells corresponding to these two nodes. We use the following condition to check whether the interaction between a pair of nodes may be used as a ``summary'' interaction:
\begin{equation}
\lVert \mathbf{y}_{cell 1} - \mathbf{y}_{cell 2} \rVert^2 / \max(r_{cell 1}, r_{cell 2}) < \rho,
\end{equation}
where $\mathbf{y}_{cell 1}$ and $\mathbf{y}_{cell 2}$ represent the center-of-mass of the two cells, $r_{cell 1}$ and $r_{cell 2}$ represent the diameter of the two cells, and $\rho$ is a speed-accuracy trade-off parameter (similar to $\theta$ in Barnes-Hut-SNE).

\begin{figure*}[h]
\centering
\includegraphics[width=.9\textwidth]{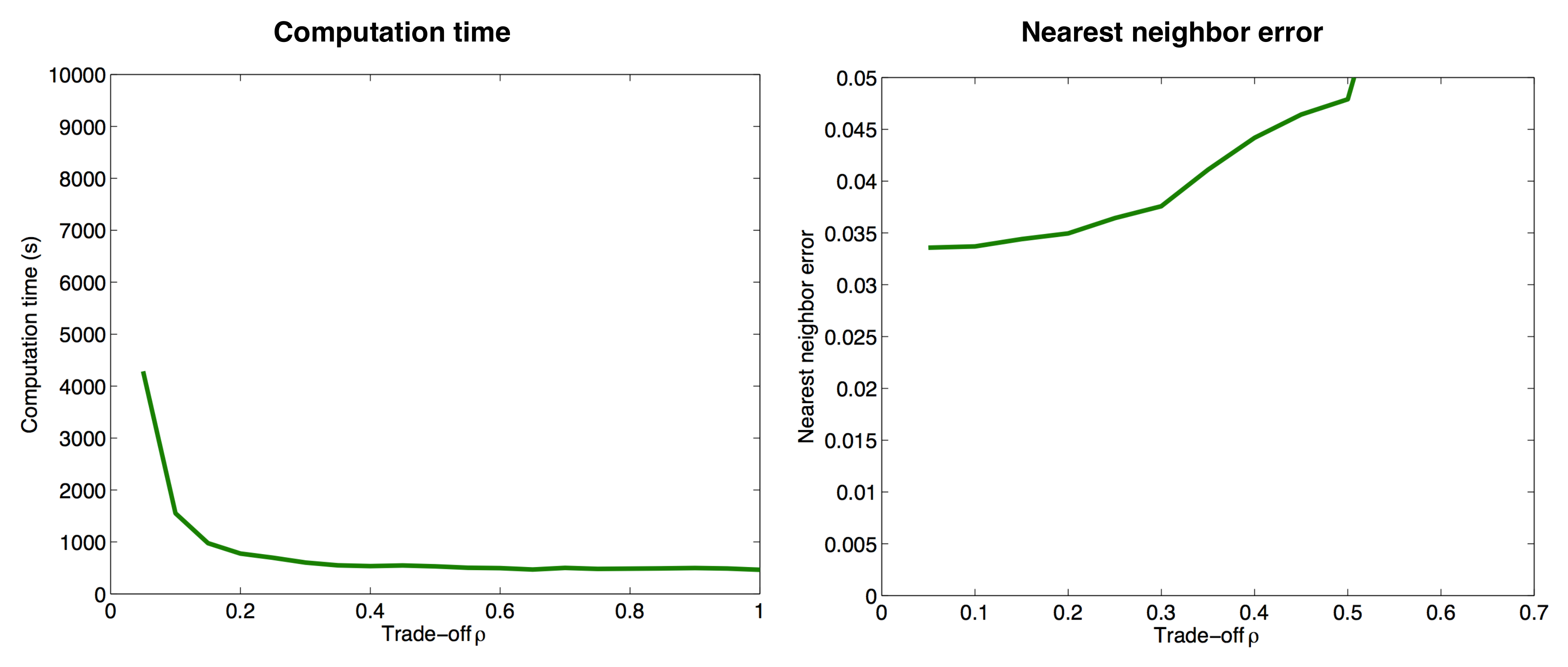}
\caption{Computation time (in seconds) required to embed $70,000$ MNIST digits using dual-tree t-SNE (left) and the $1$-nearest neighbor errors of the corresponding embeddings (right) as a function of the trade-off parameter $\rho$. The results may be compared to those in Figure~\ref{fig:mnist_theta}.}
\label{fig:app_mnist_theta}
\end{figure*}

\begin{figure*}[h]
\centering
\includegraphics[width=.9\textwidth]{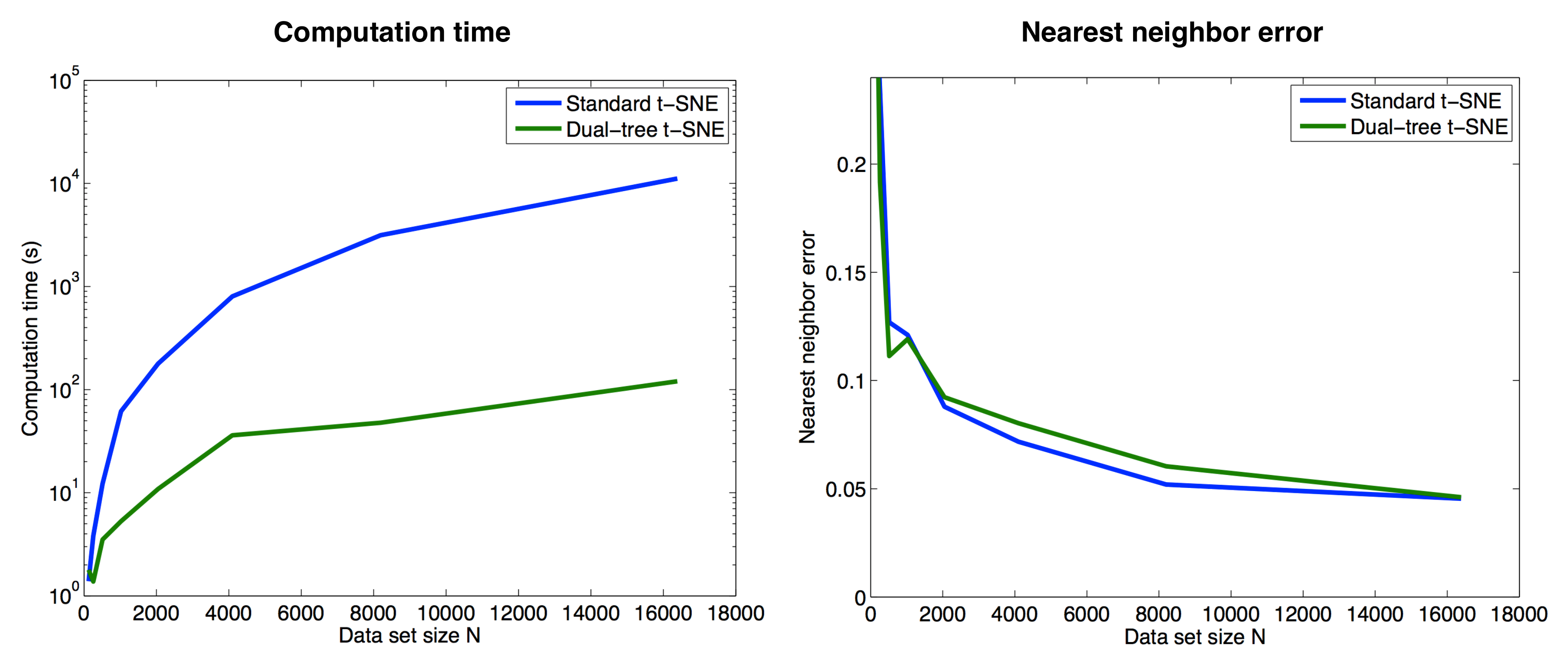}
\caption{Compution time (in seconds) required to embed MNIST digits (left) and the $1$-nearest neighbor errors of the corresponding embeddings (right) as a function of data set size $N$ for both standard t-SNE and dual-tree t-SNE. The results may be compared to those in Figure~\ref{fig:mnist_perf}.}
\label{fig:app_mnist_perf}
\end{figure*}

Figure~\ref{fig:app_mnist_theta} presents the results of an experiment in which we investigate the influence of the trade-off parameter $\rho$ on the learning time and the quality of the embedding on the MNIST data set. The results in the figure may be readily compared to those in Figure~\ref{fig:mnist_theta}. The results in the figure show that, whilst the dual-tree algorithm provides additional speed-ups compared to the Barnes-Hut algorithm, the quality of the embedding also deteriorates much faster as the trade-off parameter $\rho$ increases. The quality of the  embedding obtained with a dual-tree algorithm with $\rho\!=\!0.25$ roughly equals that of a Barnes-Hut embedding with $\theta\!=\!0.5$, and these two embeddings are constructed in roughly the same time (\emph{viz.} in approximately 650--700 seconds). Figure \ref{fig:app_mnist_perf} shows the performance of dual-tree t-SNE with $\rho\!=\!0.25$ as a function of the number of MNIST digits $N$. The results in Figure \ref{fig:app_mnist_perf} can be readily compared to those in Figure~\ref{fig:mnist_perf}. Again, the results show that dual-tree t-SNE performs roughly on par with Barnes-Hut-SNE, irrespective of the size of the data set $N$.

\end{document}